# KenSwQuAD – A Question Answering Dataset for Swahili Low Resource Language

**Barack W. Wanjawa***
University of Nairobi, Kenya, wanjawawb@gmail.com

**Lilian D.A. Wanzare**
Maseno University, Kenya

**Florence Indede**
Maseno University, Kenya

**Owen McOnyango**
Maseno University, Kenya

**Lawrence Muchemi**
University of Nairobi, Kenya

**Edward Ombui**
Africa Nazarene University, Kenya

**ABSTRACT**
The need for Question Answering datasets in low resource languages is the motivation of this research, leading to the development of Kencorpus Swahili Question Answering Dataset, KenSwQuAD. This dataset is annotated from raw story texts of Swahili low resource language, which is a predominantly spoken in Eastern African and in other parts of the world. Question Answering (QA) datasets are important for machine comprehension of natural language for tasks such as internet search and dialog systems. Machine learning systems need training data such as the gold standard Question Answering set developed in this research. The research engaged annotators to formulate QA pairs from Swahili texts collected by the Kencorpus project, a Kenyan languages corpus. The project annotated 1,445 texts from the total 2,585 texts with at least 5 QA pairs each, resulting into a final dataset of 7,526 QA pairs. A quality assurance set of 12.5% of the annotated texts confirmed that the QA pairs were all correctly annotated. A proof of concept on applying the set to the QA task confirmed that the dataset can be usable for such tasks. KenSwQuAD has also contributed to resourcing of the Swahili language.

# 1 INTRODUCTION

The quest for a question-answer (QA) dataset for natural language (NL) processing tasks continues to draw research interests globally. QA datasets are an important component in machine learning since it models one of the ways of data query that humans do. That means that machines can as well be programmed or given access to data to learn from and then undertake the same QA task for the benefit of users. QA is commonly used in many information querying tasks such as internet search, frequently asked questions (FAQ) and dialog systems.

The use of machines to process data and provide results has enabled fast information processing for the benefit of users, such as using computers for internet search. Machine processing of natural language (NL) is however not trivial. The NL has to be transformed into a format that computers understand, which could be by word embeddings e.g. one-hot encoding, term frequency inverse document frequency (TF-ID), dense vectors such as Word2vec or GloVe [1] or deep learning methods such as transformers e.g. Bidirectional Encoder Representations from Transformers (BERT) [2]. Many of these NL transformations require data for training the models and perform quite well when the training data is available in abundance e.g. BERT performance in NL tasks is well documented [3]. The data source for such machine processing tasks can be provided through a corpus of the languages being processed or through gold standard datasets for tasks such as QA.

High resource languages such as English, French, Chinese, Spanish etc. have vast amounts of data for training or even gold standard sets for QA. However, low resource languages such as Swahili, also known as Kiswahili, and many other African and world languages do not have such resources. This has been due to low research interests that would otherwise deliberately collect these datasets. The problem is made worse since those who need to do research on NL tasks are likely to have no option but to use the existing data resources that exist in high resource languages. This leads to further decline in low resource language datasets, while high resource languages continue to get more resources over time.

There is therefore a need to deliberately target low resource languages by contributing NL datasets to benefit users and researchers. The task of QA remains important for information processing by computers. The only way of ensuring that a system is capable of QA tasks is by having an existing gold standard QA set, which can then be used to verify the performance of the computer system. Deliberate research is therefore needed to get such gold standard QA sets, out of which computer processing systems that undertake QA related tasks can be tested upon.

Unfortunately, gold standard QA datasets are few or non-existent for low-resource languages. That means that research or testing of systems such as internet search and dialog systems for low resource languages is difficult. This is because QA datasets to test such models are not readily available. Contrast this to the high resource language of English where many gold standard QA datasets exits. These include SQuAD [4], MCTest [5], WikiQA [6], TREC-QA [7] and TyDiQA [8]. Very few public domain datasets have Swahili language texts, such as TyDiQA. TyDiQA is a collection of QA sets in 11 languages from Wikipedia corpus of the different languages. Other African languages are also hardly represented in standalone QA datasets. A few notable are for Amharic (Ethiopia) [9], Zulu (South Africa) [10], [11], Igbo (West Africa) [12] and Arabic language QA datasets mostly used in Northern Africa [13].

It is due to this problem of inadequacy of gold standard datasets for the low resource languages, and specifically Swahili that this research developed Kencorpus Swahili Question Answering Dataset (KenSwQuAD), a Swahili language Question Answering Dataset. We developed KenSwQuAD by annotating primary data that was collected by the Kencorpus project [40]. Kencorpus project developed a corpus of three Kenyan languages created through funding by the Lacuna fund of the Meridian Institute USA. The purpose of the Lacuna fund is to facilitate collection of data of low resource languages, both text and voice, for purposes of documenting and resourcing such languages to benefit the research community.

One of the languages in the Kencorpus dataset is Swahili, out of which KenSwQuAD was developed. Annotators formulated question and answer pairs based on over 65% of the collected texts of Swahili language. A data quality check was then done on a sample of 12.5% of the annotated data for assurance that KenSwQuAD was a correct and reliable dataset for QA tasks of the Swahili language. A proof of concept on KenSwQuAD using semantic networks and BERT-based deep learning models confirmed that NL processing was possible based on this QA dataset.

## 2 BACKGROUND

While datasets, both corpora and gold standard sets, remain important for natural language (NL) processing tasks as done by computers and on the internet, these datasets need to be developed through research for them to be realized and made available. The cycle of exploring data corpora then testing it on practical language models has led to many research efforts being concentrated on high resource languages such as English, French, Chinese, Spanish, which have readily available data corpora and language models. Testing of collected datasets using models is the sure way of confirming that the datasets are of practical use in computer processing tasks. This lack of resources (datasets, models) can discourage research in low resource languages. Deliberate research efforts for high resource languages have led to QA datasets such as MCTest [5], SQuAD [4] and TyDiQA [8]. Some low resource languages of Africa also have some QA datasets. These include Ethiopian Amharic [9], South African Zulu [10], [11], West African Igbo [12] and Arabic [13]. Such efforts need to continue for all languages, including the many more low resource languages that do not have such datasets.

It is for this reason that the quest to collect gold standard question answer datasets for the low resource languages of Africa become important. Africa has many different languages spoken within and across borders to the tune of 2,000 different languages [14]. One of these languages is Swahili, also called Kiswahili. Despite being a low resource language, Swahili is an important language of communication in Eastern Africa. It is the national language of both Kenya and Tanzania. It is also spoken in many other countries of the world. Wikipedia estimates the number of Swahili users as 200M worldwide [15], while Omniglot puts this number as 140M [16]. Swahili is therefore worthy of more research for the benefit of users and enthusiasts, both in terms of datasets and gold standard sets.

The objective of this research is therefore to develop a gold standard QA dataset for Swahili. This shall provide additional language resources to the low resource language of Swahili. The dataset is also useful for natural language processing tasks and language modeling to address issues such as internet search, dialogue systems and any machine learning system that requires question-answer type of data. The dataset shall be freely available in the public domain for research and exploration. Since the QA dataset is derived from full story texts of Swahili language, the texts themselves shall also be available for users to read and enjoy, apart from undertaking any other NL processing task that relies on raw text.

The rest of the paper is arranged as follows – section 3 provides the related work for this research, while section 4 provided the details of our methodology. Section 5 provides the results of the research, with section 6 discussing these results. Finally, section 7 provides the conclusion and points out to areas of further research.

## 3 RELATED WORK

There are many different question-answering datasets available for exploration and research. These include MCTest [5] for language comprehension questions and SQuAD [4] a dataset of 100,000 questions based on Wikipedia. Both these sets are intended for machine learning systems, which usually require lots of data for training. Lots of data resources are found for high resource languages where deliberate research collected these datasets over time.

High resource languages have benefitted from the existence of many data sources and hence machine learning models have been developed to exploit datasets. For example, Wikipedia [17] has been used as the data source of machine learning based QA systems due to the vast data available on that site. Even SQuAD is a collection of QAs from Wikipedia. Models based on SQuAD also use Wikipedia for their training.

Low resource languages have however received less attention in the development of datasets and machine learning models [18], [19]. Question answer datasets such as TyDiQA [8] have QA sets in more languages than just the high resource ones. TyDiQA is a collection of QA sets in 11 languages, both high resource and low resource languages. It is one of the few sets that has the Swahili low resource language as part of the collection. It is also based on Wikipedia articles and the QA pairs are crowdsourced from web users. It exploits this vast data source and hence provides data that can be used for machine learning. However, even the TyDiQA dataset for Swahili has deficiencies such as incorrect responses to some questions as per analysis done in other research [20]. This could be due to the crowdsourcing nature of the dataset. It however remains among the few sets that deliberately setups up QA pairs for some low resource languages and makes it available in the public domain.

Datasets that target low resource languages remain comparatively few, yet machine learning models require such training datasets. This deficiency has continued to disadvantage the low resource languages since models continue being tested and improved for high resource languages, such as English. This has progressively led to continued neglect of research efforts in resource low resource languages [21]. Some efforts have nonetheless been made to uplift the resources available for low resourced languages. These include the Helsinki corpus of Swahili [22] and Swahili language online part of speech tagging tool Swatag [23]. The Kenyan Kikuyu language has a spellchecker [24], while a named entity recognition set for ten languages of Africa has been developed [25].

Some QA datasets have been developed for African languages, though few are available for languages in the East African region. A factoid [26] [27] and non-factoid [28] QA datasets for Amharic language, mainly spoken in Ethiopia, have been developed. The data collection method used by the non-factoid QAs is by human annotators using data from websites. This QA dataset has 1,500 questions. Another dataset for factoid questions for Tigrigna language of Ethiopia, based on newspapers as a data source, has also been developed [29], though the details of QA formulation are not clearly provided. Other researchers have developed a speech QA system of five languages, including Swahili [30], to augment TyDi-QA [8] with a speech component. This speech QA has 68k audio prompts from 255 annotators. The audio data is collected using mobile phone apps. The dataset is used for Automatic Speech Recognition (ASR) tasks, minimal answer applications, and QA tasks.

Other available resources include a proof-of-concept speech QA system for the Zulu language, spoken predominantly in South Africa, though for a small experimental set [10]. Another study collected 230K domain QA pairs from medical records for purposes of automating a digital helpdesk for South African languages including Zulu, Afrikaans, and Xhosa [11]. In the West of Africa, some tests have been done on the QA task for Igbo, Hausa, and Yoruba languages. The data source was an initial English QA dataset of 183 QA pairs, which was then translated into the respective languages [12]. The Arabic language predominantly used in Northern Africa has several QA systems such as TREC [31], DAWQAS [13], TALAA-AFAQ [32] and QA4MRE [33], with DAWQAS having the largest set of 3,205 QA pairs.

The observable trend in these QA systems is their general small size in terms of number of QA pairs and their limited representation of the otherwise many languages that are spoken in Africa. These QA systems show a good start in the development of similar datasets for African languages. More research is needed to resource these and many of the other existing low resource languages of African and the rest of the world.

The starting point of resourcing low resource languages remains the provision of more datasets, tools, and gold standard datasets such as QA sets. Testing of such datasets for low resource languages may however not be done using machine learning methods that need data for training, hence other methods such as language modeling using semantic networks (SN) can be explored [34], [35]. Such modeling does not need training data but still presents the data in a way that machines can read, understand, and then perform practical tasks such as question answering. SNs are already being used in domains such as Google Knowledge Graph [36], LinkedIn [37] and Facebook [38] amongst others. Deep learning methods, such as Bidirectional Encoder Representations from Transformers (BERT)-based systems have shown very good performance for high resource level language applications when exposed to large volumes of training data. Their performance in low-resource languages is however not as good.

Low resource languages therefore need to be given research focus by providing data sources and datasets [39]. Additionally, we already have data modelling methods such as semantic network representation that do not need training data to represent such datasets in ways that make them of immediate benefit to users e.g. for internet search or even question answering. However, in time the low resource languages shall be resourced and even their processing may as well be done using the better performing statistical methods that need training data as already proven with high resource languages. Before then, deliberate effort of developing the datasets should continue. This realization is what informs the need for a gold standard QA dataset for Swahili as done in this research.

## 4 METHODOLOGY

Kencorpus Swahili Question Answer Dataset, KenSwQuAD, was formulated using a method comparable to that used for SQuAD [4] and TyDiQA [8] but was tweaked to suit the available data source. The dataset used for generating KenSwQuAD was the Swahili portion of the data collected by the Kencorpus project [40]. Kencorpus project collected primary data, both text and voice, in three low resource languages of Swahili, Dholuo and Luhya. The first language listed, Swahili, also known as Kiswahili, is spoken throughout East Africa and in other parts of the world, while the last two languages are predominantly spoken in the western part of Kenya, and parts of Uganda and Tanzania that neighbour these populations. The Kencorpus dataset contains both the language data, and the support datasets such as part of speech (POS) tagging and translations between the languages (Dholuo-Swahili and Luhya-Swahili). The corpus and datasets were developed for machine learning applications. The project did quality control checks on data collection and annotation by the engagement of language and subject matter experts at the various stages of the project.

Kencorpus project collected primary data from institutions of learning and from the speakers in the local Kenyan communities where the three languages are used. Researchers, who were natives of the languages, did direction interaction with the data sources for purposes of the data collection. Written texts from primary sources were mainly derived from story writing competitions in schools. These texts were mostly handwritten when collected from schools or already in computer format when collected from colleges. These handwritten texts were then retyped or processed using optical character recognition (OCR) software into computer formats. The project also collected secondary data from publishers and media houses. These were already typeset texts either directly in computer format or provided as hard copies for eventual conversion to computer format.

All collected text data was then converted to a final computer format (text, TXT format), was then finally cleaned up by researchers who were also natives of these languages, into a final text for inclusion in Kencorpus project dataset. Each text was collected alongside its metadata to describe aspects such as description of source, dates, geographical locations, original, preprocessed, and final formats etc. The story genres in the Kencorpus collection were varied, including news articles, book sections, discussions on topical issues, folk stories, and current affairs.

### 4.1 Data selection

The Swahili dataset in Kencorpus had 2,233 unique texts and 104 unique voice files. KenSwQuAD used purposive sampling to shortlist some of the Kencorpus Swahili texts for QA annotation. The research adopted this sampling method because this was the first time such a dataset was being developed. The researchers needed to develop a predefined criteria for the choice of texts that met the annotation objective. The project time and funding were also limited, hence the research had to get the balance right in the data selection stage to realize the deliverables within these constraints.

The shortlisted texts were those that conformed to predefined criteria. For example, the selected texts were to be at least 100 words in length but not more than 2,000 words. This was done to eliminate very short or very long texts that would have been difficult to annotate or to follow. The shortlist also targeted only prose and short stories. The research had realized that annotating texts such as plays, dialogue or poems would be difficult for the QA task since our research was based on the methodology adopted by SQuAD [4] and TyDiQA [8]. Due to this consideration, items such as poems, Tweets, Facebook posts, long stories, religious texts, text on comics, texts of mixed languages and songs were excluded from the shortlist. The summary of the selection criteria is shown in Table 4.1 below. The first column of the table describes the aspects considered in our QA dataset. We started with 2,585 candidate Swahili texts in the corpus. We then reviewed this full set of 2,585 texts and excluded the texts that did not meet our set criteria. For example, we excluded 42 texts that had more than 2,000 words.

Table 4.1: Summary of texts sampled from the Kencorpus

| Consideration | Total No. |
| --- | --- |
| Total Swahili texts in corpus | 2,585 |
| Excluded texts with over 2,000 words | -42 |
| Excluded texts with under 100 words | -325 |
| Excluded texts for any other reasons | -50 |
| Total candidate Texts shortlisted for annotation | 2,168 |
| Total text provided to annotators | 1,660 |
| Proportion provided for annotation | 76.6% |

The final set of shortlisted Swahili texts based on the selection criteria, were therefore 2,168 texts, of which 1,660 texts (76.6%) were provided to the QA annotators. The method of data allocation to annotators was by equal number of texts in each time duration (monthly, at the start of the month), then allowing individual annotators to access the next set of a fixed number of texts upon finalization of their targets. These subsequent sets were allocated weekly and replenished weekly upon confirmed completion. All the work was done in the 2 months total project duration.

### 4.2 Annotation guide

The research developed an annotation guide that spelt out the expectations of the annotation project, including issues such as the inclusion and exclusion criteria. This was done to assist the researchers to pick the right story texts from the vast Swahili dataset from the Kencorpus project datasets. We availed a final shortlist of 1,660 texts to the annotators out of the initial candidate dataset of 2,168 texts. After the selection of texts that met the annotation criteria, the research developed the criteria for the number of questions, types of questions and the type of answers to annotate. An analysis of the questions set on the TyDiQA [8] dataset informed the type of questions to formulate for KenSwQuAD.

The research decided to set a standard number of five questions per text. The questions were to be the object enquiry type (what, which, who, when). The research also decided to include at least one question that involved some reasoning (why, how). Most QA datasets usually revolve around such enquiry, including SQuAD (though it is in English language) and TyDiQA (the Swahili portion). These guidelines were also included in the annotation guide.

Additionally, the research provided guidance to the annotators on the number of questions to set for each question type, the desirable number of words in the question text and the answer text. Guidance was also provided on whether unanswerable questions should be allowed. The issues considered are summarized in Table 4.2 below, as was also included in the annotation guide. Some of the provisions in the guide were just suggestions e.g. number of words or questions to be set per type. The final decision depended on the text of the story, though the annotators were asked to try their best to follow the recommendations in the guide. Column one of Table 4.2 describes the various considerations done when setting up the Q-A pairs. For example, we recommended that 3 out of the 5 questions annotated on any text should be the who/what/which type of question, which we called 'Type 1' questions. Enquiry on temporal issues (when) and descriptive aspects (how or why) were limited to one annotation each. Other aspects considered were the number of words in the question or answer. These considerations were to enable us to have generally uniform types of annotated Q-A pairs, despite the annotations having been done by different annotators.

**Table 4.2**: KenSwQuAD criteria for setting QAs

| Aspect | Recommendation |
|---|---|
| Number of Type 1 questions (who, what, which) | 3 |
| Number of Type 2 questions (when) | 1 |
| Number of Type 3 questions (how, why) | 1 |
| Number of questions per story text | 5 |
| Number of words in the question | 10 max |
| Number of answers per question | 1 |
| Number of words in the answer (Type 1 and 2) | 3 max |
| Number of words in the answer (Type 3) | 10 max |
| Multiple choice answers permitted | No |
| Unanswerable questions permitted | No |

### 4.3 Annotators selection

The research recruited annotators then conducted both in person and online training for the annotators. The project was done during the global Coronavirus disease (COVID-19) pandemic (2021- 22) when social distancing was necessary and in-person meetings were limited. The recruitment of annotators targeted speakers of Swahili language and additionally those who were currently engaged in the teaching of the Kiswahili language in Kenyan educational institutions. This requirement was set to give the project the best personnel who could understand the intricacies of question and answering formulation based on their own experiences with the Swahili language in their careers. QA sets also tend to mimic the real-world information retrieval needs; hence the annotators would leverage on their experience on what learners would usually query on such a corpus of story texts. They would also consider what they would normally examine as teachers while testing language comprehension. Nonetheless, the research restricted the type and complexity of questions and excluded issues such as questions on language structure e.g. parts of speech, role of words etc.

The in-person training was done over a two-day workshop where the annotation guide was discussed, and practical demonstrations done. This was followed by presentations, discussions, and consensus building. The review and evaluation of the workshop confirmed that the annotators understood the expectations of the project. Online training was done for both those who had attended the physical workshop and to the new members who had met the recruitment criteria and were joining the annotation team. The new members were assigned to respective mentors who had attended the in-person workshop. A practical training on annotation was once again done during the online meeting and discussions held on dos and don'ts of the annotation project.

### 4.4 Annotation tool
The researchers developed an online annotation tool on Google forms for collecting the QA pairs. The annotators were trained in the use of the online form and did a practical on the use of the tool during the training sessions. The annotation team of six was then given twenty sample story texts each and asked to test the annotation tool and give feedback on their experiences over a period of one week. The tool was tested while the researchers monitored the data that was trickling in on the backend collation spreadsheet. The annotators confirmed that they were conversant and ready to use the tool based on their one-week test period. The submitted test annotations were carefully reviewed by the researchers to confirm the annotators work. We also confirmed that the QA pairs were fit for the project as per the annotation guide, training, and discussions.

**4.5 Data size and Distribution to annotators**

We reset the data collection database after the test period and provided the annotation team with a new set of texts for actual annotation. Each annotator was allocated different and unique texts to annotate, still based on purposive sampling. The sampling was done as follows – first, the collection of 1,660 texts was split into the format of their raw data e.g. those already typed texts in computer format (TXT, DOC, RDF) and images (PDF, JPG, PNG). The images were further categorized into handwritten versus those from typed sources.

The Kencorpus metadata had already provided information on the exact or approximate number of words on each text, hence each of the categories of texts was then sorted by the number of words. Each of the six annotators was then allocated the texts in a category. This was done by allocating a unique story text to one annotator at a time, then the allocation was repeated in batches of six until all the texts in the category were exhausted. The allocation of texts to the six annotators would then be done in the next category of texts as per the list of texts as sorted by number of words. This sampling ensured that each annotator got exposed to all types of texts (as per the different text types and formats) and also got comparable lengths of texts in an equitable manner.

We also spelt out the output expected per week, both in terms of minimums and maximums, just to ensure that the annotators gave the project the expected concentration that was needed. Doing too much work over a short period of time had the danger of the annotators rushing through their work. This would likely mean that they were not giving each text the time of reading and thoughtfulness that was needed to ensure that they formulated the expected quality of QAs.

These measures were monitored weekly over the two-month annotation period. The distribution of data to annotators is shown on Table 4.3 below. Column one of Table 4.3 shows the volume of texts that we considered, starting from the full corpus to our final dataset. We started with 2,585 total number of Swahili texts in the Kencorpus project, then used our shortlisting criteria to select 2,168 candidate texts for the QA annotation task. Out of these, we only managed to provide 1,660 texts for the QA annotation task. In anticipation for future machine learning purposes and to further verify the exact locations in the text where the answer can be found, we also annotated some texts with paragraph numbers. We therefore deliberately provided a 13.1% set of texts (218 texts), out of the total 1,660 texts, to the annotators to include the paragraph number from where they got their answers from.

**Table 4.3**: Description of Data provided to annotators of KenSwQuAD

| Aspect | No. | Comment |
|---|---|---|
| Total no. of texts in Kencorpus Swahili | 2,585 | All texts in corpus |
| Total no. of texts phortlisted for KenSwQuAD set | 2,168 | 83.9% of text corpus |
| Total no. of texts provided to the annotators | 1,660 | 76.6% of shortlisted |
| Total no. of texts to annotate without indicating paragraph | 1,442 | 86.9% of provided data |
| Total no. of texts to annotate and indicating paragraph no. | 218 | 13.1% of provided data |

**4.6 Quality control check on annotated data**

We finally did a quality control check on the annotation work. This was done over a one-week period at the end of the annotation project, where we sampled 12.5% of the texts already annotated (180 texts with 900 questions) and switched them through to different annotators, ensuring that none of the annotators got their own work.

The switched over dataset consisted of the unique story text identification numbers (story_ID) and only the questions that had been set by the initial annotator. We deleted the original answer and left that answer column blank. This annotation set was provided to the annotators in the form of a spreadsheet. We then provided each annotator with the relevant texts and the spreadsheet containing the questions only. Their task was then to derive only the answers after reading the text and the associated questions. The 180 texts sampled had both the QA types where paragraph number had been indicated (13%), while the balance 87% did not have paragraph numbers indicated. This sampling ensured that both types of QA annotation types were selected in the same proportion as their numbers in the original set as we did this quality control check.

### 4.7 Proof of concept testing of the QA dataset

As proof of concept, the research used the semantic network (SN) method to create some networks of the final QA dataset texts to check if such an SN machine learning method was capable of undertaking QA tasks. The methodology used for setting up the model was that already done for Swahili QA in previous research [20]. This method works in cases where there is no data for training a model. In this method, the Swahili text is first subjected to part of speech tagging (POST) which can be done using online tools [23], followed by the creation of an SN connecting subject-predicate-objects (SPO) of the text using tools and programming code. The created SN is then queried using a query language - SPARQL Protocol and RDF Query Language (SPARQL). We also tested a deep learning method based on transformers - Bidirectional Encoder Representations from Transformer (BERT) with multi-lingual pretrained data that has some Swahili data i.e. Robustly optimized BERT approach (RoBERTa), specifically, XLM-RoBERTa [41]. However, new transformer models continue to be developed and one that is specifically tuned for Swahili [42] is now available. We did not have an opportunity to test it on QA. So far other researchers [42] have tested it on the tasks of classification and named entity recognition (NER).

## 5 RESULTS

The result of the Swahili language QA annotation project is the Kencorpus Swahili Question Answer Dataset, KenSwQuAD. This started from the initial identification of 2,168 unique texts that were shortlisted for the project, out of which 1,660 of these texts were provided to the six annotators.

### 5.1 KenSwQuAD corpus statistics

The QA annotation responses from annotators was recorded on an online form. At the close of the project, a listing of 1,547 entries had been posted on the online collation datasheet. However, only 1,370 texts had been uniquely numbered as expected, to correspond to their respective story_IDs as provided to QA annotators. The balance of 177 texts listed in the data collection tool had repeated story_IDs. The analysis of the 177 annotations with repeated story_IDs is shown on Table 5.1 below. The column labelled 'Set' shows the total number of texts in that category, 'Action' indicates what needed to be done on that set, while the column 'Final' shows the total number of texts with unique story_IDs that were reconfirmed after reverification of the QA annotations. For example, our analysis confirmed that 26 texts had instances of repeated story_ID numbers. Sets of texts turned out to be exact duplications, meaning that the annotators posted the QA pairs twice on our annotation tool. We therefore only picked 12 unique story_IDs out of this set, each with 5 QA annotations. An analysis done on another set of 129 instances of repeated story_IDs led to a final identification of 63 unique texts. Different QA pairs had in some instances been annotated for the same story text, hence we ended up with more than the expected 5 QA pairs per text from this set of 63 texts.

**Table 5.1**: Analysis of Repeated Stories collected during annotation

| Aspect | Set | Action | Final |
| --- | --- | --- | --- |
| Total no. of repeats of story_IDs (exact duplication) | 26 | Pick one only | 12 |
| Total no. of repeats of story_IDs (different QAs set for the same story_ID) | 129 | Combine the collected QAs to be more than 5 sets per story_ID | 63 |
| Total no. of annotations with repeats of story_IDs (different QA different QA contexts) | 22 | exclude from collection since story_IDs are erroneous | 0 |
| Total | 177 | | 75 |

From the results shown on Table 5.1, the reconciliation of annotations from repeated story_ID numberings resulted into 75 additional texts with uniquely identified story_IDs. The 75 unique texts were therefore added to the initial collection of 1,370 texts to give a final total of 1,445 texts, each annotated with at least 5 QA pairs. This final set of 1,445 unique texts with QA pairs was derived from both the set of those texts that had paragraph numbers indicated (201 texts) and those that did not indicate the paragraph number (1,244 texts). The derivation of the final KenSwQuAD dataset is shown on Table 5.2 below. The table shows the progression of data numbers from the initial Kencorpus project collection of 2,585 texts to the shortlisting of only 2,168 that met our inclusion criteria for the QA annotation. Though we provided 1,660 texts for QA annotation, the initial result from the data collection tool listed 1,547 entries as the number of annotated texts. However, further analysis of the 1,547 listed texts identified 1,445 texts as the final number of unique annotated texts. This is the final number of texts that we compiled in the final KenSwQuAD QA dataset. KenSwQuAD dataset therefore consists of 1,445 unique story texts each annotated with a minimum of 5 QA pairs, giving a total corpus of 7,526 QA pairs.

**Table 5.2**: Results of annotating KenSwQuAD dataset

| Aspect | No. | Comment |
|---|---|---|
| Total no. of texts in Kencorpus Swahili | 2,585 | All texts in corpus |
| Total no. of texts Shortlisted for KenSwQuAD set | 2,168 | 83.9% of text corpus |
| Total no. of texts Provided to the annotators | 1,660 | 76.6% of shortlisted |
| Total no. of annotations collected on the online form | 1,547 | Not all were unique |
| Total no. of texts with unique story_IDs for QA set | 1,445 | 87.0% of the provided texts |
| Total no. QA pairs set on the unique story_IDs | 7,526 | At least 5QAs per text |
| Total no. of texts that were not annotated | 215 | 12.9% of provided |

**5.2 KenSwQuAD quality control check**

The results of the quality control check done on the sampled data from the collected QA dataset is shown on Table 5.3 below. Column one on Table 5.3 breaks down the salient features of the quality checks done on the KenSwQuAD QA dataset that had 1,445 unique texts. The quality checks were done on a set of 180 texts (12.5% of the total) from the QA dataset, consisting of 900 QA pairs. In 485 out of 900 questions (53.9%), the answers provided by the second annotators were exact word-for-word match with the responses provided by the first annotator. However, in another 37.9% of the cases, some answers were not the exact words used by the original annotators, but the context or reasoning was the same e.g. 'Mama yake' and 'mamake' (his/her mother) or 'Miaka 25' and '25' ('25 years old' versus '25' as a number). These were deemed to be the same since they would naturally be answered in either of those two versions. Other cases where the original and second annotator were deemed to have agreement was when there was an obvious typo during the data entry by either of the two annotators (3.8% of the cases). The analysis of this quality control evaluation is shown on Table 5.3 below.

**Table 5.3**: Results of quality control check on some KenSwQuAD texts

| Aspect | No. | Proportion |
|---|---|---|
| Total no. of texts finally annotated for KenSwQuAD | 1,445 | 100.0% |
| Total no. of texts sampled for quality control check | 180 | 12.5% |
| Total no. of questions in the check dataset | 900 | 100.0% |
| Total no. of answers that were as exact word-for-word as initial annotation | 485 | 53.9% |
| Total no. of answers that were as similar to initial annotation | 341 | 37.9% |
| Total no. of answers that were as deemed comparable | 34 | 3.8% |
| Total no. of answers that were not as per initial annotation | 40 | 4.4% |

The question that had the least agreement between annotators was Q5, that required some reasoning. The general thought of the annotators was the same, but the words used in the answer set were quite different. We considered the answers to be similar in cases where the same reasoning (not necessarily the words) was used in the how/why type of questions e.g. 'name one of the characters' would be correct for any qualifying answer, even if the annotators provided different answers.

As indicated on Table 5.3, there was no agreement between the original and the checking annotators in 40 QA pairs. We analyzed this set of 40 cases as per Table 5.4 below. The final correct or wrong verdict was made by the third reviewer, being the researcher. In this set, we had 9 instances where the initial annotator was correct, while the reviewing annotator was wrong, while in one instance the initial annotator provided a wrong answer though the reviewing annotator provided the right answer. This one wrong case where the initial annotator was considered incorrect was however not completely out of context. The initial annotator indicated the answer as 'tree' (that bears the fruit), while the second annotator indicated the answer as 'fruit' itself. The context of the question would however strictly restrict the answer to 'fruit'

**Table 5.4**: Analysis of cases of disagreement between annotators for KenSwQuAD answers

| Aspect | No. | Proportion |
|---|---|---|
| Total no. of answers with disagreement | 40 | 100.0% |
| Total no. of answers where initial annotator was correct | 9 | 22.5% |
| Total no. of answers where initial annotator was wrong | 1 | 2.5% |
| Total no. of answers where both initial and reviewing annotators were correct upon review | 30 | 75.0% |

The analysis done on this set of disagreements also confirmed that in 30 out of the total 40 cases (75% of the cases), both the initial and reviewing annotators were correct, but did not use the exact same wordings in their answers. These cases were mostly observed in the question 5, which was a how/why question.

**5.3 KenSwQuAD proof of concept on QA task**

For proof of concept on a machine learning task, some texts from the QA dataset, such as story_ID 3830 (five paragraphs with 354 words) was converted to a semantic network (SN). The partial reproduction (paragraph 1 only, 58 words) of the text in Swahili is shown on Table 5.5 below. The translation of the text to English is also provided for purposes of illustration.

**Table 5.5**: Original and Translation of story_ID 3830 from KenSwQuAD dataset

| Source text | Sample Content of Story text |
|---|---|
| Original (Swahili) | *Kilimo katika nchi yetu ya Kenya ni muhimu na kinafaa kuzingatiwa kwa manufaa yake mengi. Moja ni ufugaji wa mifugo ambao hutupa protini kupitia kwa nyama. Hii protini ndiyo inayoupa nguvu mwili na kuujenga. Mifugo hawa kama kuku hutupa mayai ambayo yaweza yakauzwa na kuimarisha maisha ya mfugaji na familia yake kwa kumpa pesa za kukidhi mahitaji yake* |
| Translation (English) | Agriculture is important for Kenya hence needs attention due to the many benefits. First benefit is the protein obtained from meat. This protein provides energy to our body and builds it up. Animals such as chicken provides us with eggs which can be sold to benefit the keepers and their families by providing them with cash that is used for their wellbeing |

A partial semantic network created from story_ID 3830, part of the text shown in Table 5.5, is shown in Fig. 5.1 below. The figure is the visual representation of the resource description framework (RDF) formatted file as produced using online tools [43]. The interrelationships between the various subjects and objects can be visualized in the figure, as contained on the RDF triples data store.

The 5 questions annotated under KenSwQuAD and subjected to the network are shown on Table 5.6, with a translation to English indicated in brackets for purposes of illustration. The 5 questions formulated for the story_ID 3830 are shown on the second column of the table. The gold standard answers provided by the annotators are also provided on the last column.

**Table 5.6**: QA set for KenSwQuAD story_ID 3830

| QNo. | Question | Answer |
|---|---|---|
| 1 | *Kilimo ni muhimu katika nchi gani* (In which country is agriculture important) | Kenya (Kenya) |
| 2 | *Mifugo hutupa nini* (What do animals provide) | *Mbolea* (fertilizer) |
| 3 | *Mahindi huuzwa wapi* (Where is maize sold) | *Ng'ambo na nchini* (Both within the country and abroad) |
| 4 | *Ni aslimia gani ya vifaa vinavyouzwa ng'ambo* (What percentage of good are sold out abroad) | 80 (80%) |
| 5 | *Vipi maisha huweza kuimarishwa kupitia kwa kuku* (How do chicken improve the wellbeing of keepers) | *Kwa kuuza mayai* (By the sale of eggs) |

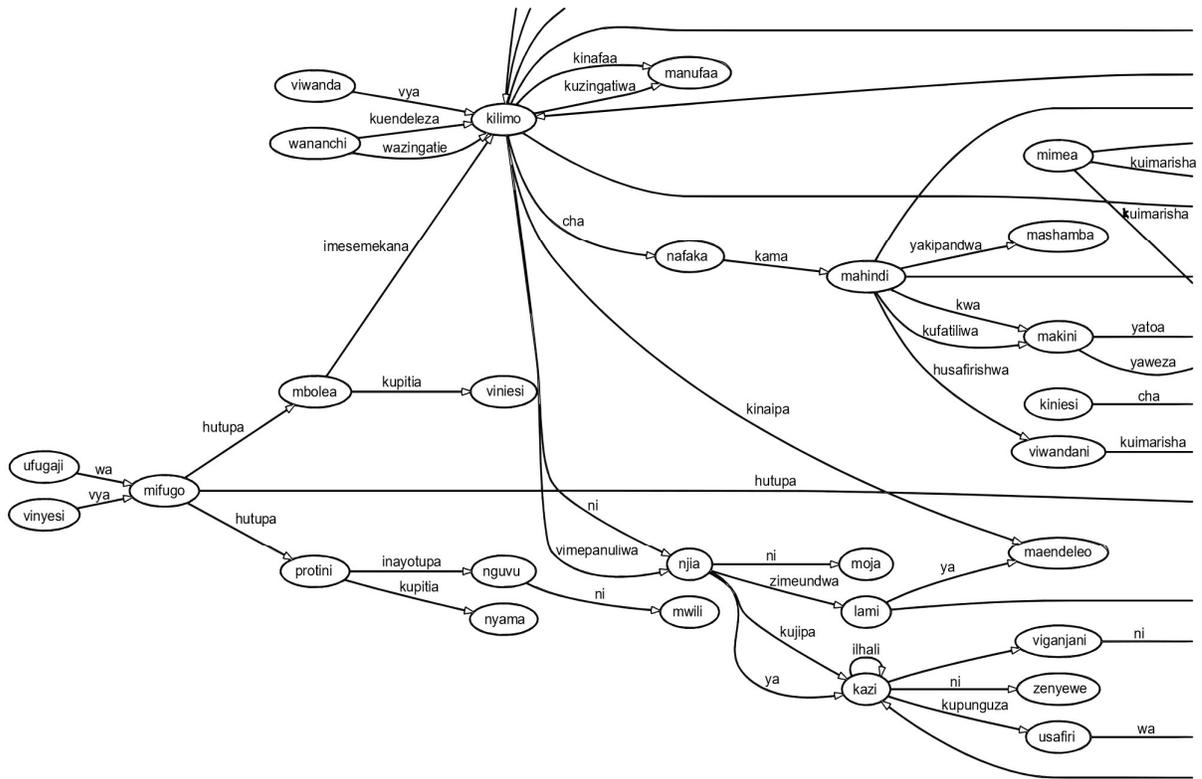

**Fig. 5.1**: Visualization of RDF of the Swahili language text created from Kencorpus story_ID 3830 (source: author)

We subjected the semantic network system to the same set of questions and compared the generated answers to the gold standard answers. Carefully formatted query language (SPARQL) queries subjected to the network to enquire on the subjects and objects could easily provide responses to questions 1, 2 and 4 based on the network. The SN answered Q3 as 'ng'ambo' (abroad) and 'Kenya', since that is what is explicitly stated in the text. The response of the annotator was however 'abroad' and 'within the country'. Of course, by implication, 'within the country' would mean 'Kenya', but that is not explicit from the annotator's response. This is an issue that differentiates the explicitness that such computer algorithms require, versus the generalization/assumptions of human reasoning.

The object enquiry questions (1,2,3,4) were therefore all answered correctly based on the SN, with only the reasoning question being unanswerable (Q5/A5). Formulating a SPAQRL query to address this type of question is difficult as a start, and the nearest that could be done was to ask for a relationship between 'chicken' and 'livelihood'. This relationship gave a different result 'eggs', and not 'sale of eggs', which was the monetary value derived from this transaction, as indicated, and required by the annotator. This was always the intent of Q5 of all the annotated QA pairs – a reasoning question that should not be directly picked from the text but needed some reasoning, despite the facts on the text. The proof done in this research was limited to only a few texts, since the performance of this method has already been tried elsewhere [20]. This method is shown to vary in accuracy depending on the complexity and size/words on the text being formatted as an SN. It performs well on short texts where the network tree is not so complex.

This research also did another proof based on BERT. We did not find any specific monolingual language model for Swahili, hence had to use multilingual BERT, specifically XLM-RoBERTa base model [41]. We exposed the pretrained model to a dataset of contexts with 800 QA pairs for training and another set of contexts with 400 QA pairs for testing. The settings used in our experiments were the defaults, but with specific changes such as batch size = 12, Learning rate = 3e-5, Epochs = 40,

Max sequence length = 384 and max split span = 128. Input data files were formatted as JavaScript Object Notation (JSON) files. We created Python scripts and files as guided by XLM-RoBERTa documentation [44] and used Google Colab online runtime environment with GPU instance. The training and testing on this setting took about 1 hour.

We obtained an F1-score of 59.4% and an exact match score of 48.0%. These results are not as high as those obtained when using English language datasets, such as SQuAD, on the task of QA, where BERT-based ensembles have achieved F1-scores of up to 93.214 [45]. The semantic network-based models used in the proof-of-concept performed better than the BERT-based model. The performance bottleneck is likely due to the lack of vast amounts of training datasets for the low resource language of Swahili. However, these results from BERT-based model confirm that the KenSwQuAD dataset is capable of use in the machine learning task in QA. The comparison of the two proof-of-concept models is shown on Table 5.7 below.

**Table 5.7**: Comparing Machine learning models using KenSwQuAD dataset

| Model | Performance | Metric |
| --- | --- | --- |
| Semantic Network based | 80% | Precision@1 |
| BERT-based | 59.4% | F1 score |

### 5.4 Final KenSwQuAD corpus

The result of this project is the Kencorpus Swahili Question Answer Dataset, KenSwQuAD. This is a dataset of 1,445 story texts, each annotated with at least 5 QA sets, hence a total of 7,526 QA pairs. The texts are provided in a corpus while the QA set is provided as a comma separated values (CSV) file format. Both the dataset of context texts and the annotated QA pairs are available in CSV format and have been released to the public domain on either of the following data repositories: **kencorpus.maseno.ac.ke/kenswquad https://doi.org/10.7910/DVN/OTL0LM**

### 6 DISCUSSIONS

The results of this research show that it is possible to develop a question answering dataset for a low resource language such as Swahili, the product of which is Kencorpus Swahili Question Answer Dataset, KenSwQuAD. This is possible when the data of the language is collected then an annotation guide is developed to guide the QA annotation process. Checking on the work being done by the annotators through continued monitoring ensures that the expectations of the annotation are achieved. Those who use and understand the language should ideally be selected to do the annotation for purposes of getting the best gold standard set. However, it is essential to check on the quality of the work done, as was done, by sampling the annotated work and asking for different annotators to reconfirm the answers.

The KenSwQuAD set managed to annotate 7,526 QA pairs from 1,445 texts by annotating a standard number of 5 questions per text. This set was developed from 55.9% of the whole collection of 2,585 Swahili texts in the Kencorpus data collection. However, the sampling criteria that we adopted meant that only 2,168 texts of the 2,585 were shortlisted for annotation, with 76.6% (1,660 texts) from the shortlisted texts being availed to the QA annotators. The annotators therefore managed to set QA pairs for 87.0% of the texts (1,445 texts) provided to them for annotation. In terms of the annotation work itself, the annotators were able to work on most of the texts that they were allocated. The texts that were skipped were those that had issues such as being illegible or those that were not making sense based on what was written by the source authors. Some annotators were however not able to finish the batch of story texts that they had been allocated by the end of project. However, all annotators managed to achieve at least 80% of their targets. This shows that careful selection of annotators is essential to achieve QA project objectives such as that of KenSwQuAD.

A quality control set of 12.5% of the texts in the final annotated corpus confirmed that the initial annotators had done the correct annotations, since only 1 of the 900 tested QA pairs was incorrect. An analysis of the incorrect response nonetheless showed that it was still within context. However, getting exact words in the answer from both annotators was only possible in 54% of the cases, mostly for object or one-word responses. The rest of the answers had a slight variation e.g. in spelling, word inflections or formation, use of synonyms or use of phrases that meant the same thing. The KenSwQuAD QA set is therefore reliable and is confirmed to contain questions that would elicit the expected responses regardless of the annotator. Ambiguity in language is however an issue to contend with when annotating QA pairs.

There are many different types and range of QA systems, and one QA annotation may not cover all aspects of QA. It is for that reason that the project developed an annotation guide that set the boundaries of our dataset and spelt out what could or could not be done. For example, our QA annotation did not cover aspects such as the unanswerable questions or the cloze type of QAs. Such restrictiveness is also true for most public domain QA systems which usually pick a particular type of QA setup and stick with that type only. This research followed suit by annotating the machine reading comprehension (MRC) type of Swahili QA.

The KenSwQuAD set is therefore a good starting point for the different natural language tasks that require question-answer sets. These include QA systems, internet search and dialogue systems. This dataset can be a starting point for such systems. Machine learning of Swahili and other low resource languages shall start being more prevalent with continued data collection and development of similar QA annotations as KenSwQuAD. This shall lead to more interests in such languages, which is likely to lead to their resourcing e.g. by more datasets and more model testing.

A proof of concept on the applicability of KenSwQuAD was done by trying the data on a machine learning system of semantic network (SN) generation from annotated text of the corpus. The SN method tested was that used in other related research [20] that had been proven to work in cases where there is no data to train a model. This was previously tested using the Swahili part of the TyDiQA dataset [8]. The results of testing the KenSwQuAD dataset confirm that the natural language text can be formatted as an SN. The results of querying the SN as evaluated against the gold standard QA pairs of KenSwQuAD confirmed that the expected answers were generated. This research demonstrated that the use of SN can format the data in such a way that computer systems can understand this low resource language of Swahili and answer some questions correctly.

Another proof done using transformer models (XLM-RoBERTa) achieved average performance, though this was lower than the performance of the SN-based model. This can be attributed to the relatively low volume of training data for the low-resource language of Swahili in the multi-lingual BERT model. However, even this average performance just shows that machine learning systems that need training data are still capable of processing Swahili language as they progressively get exposed to datasets such as KenSwQuAD.

The KenSwQuAD dataset is therefore of benefit to the computer science and machine learning community who want to build models that understand language and then undertake useful tasks such as internet search, dialog systems and chat bots. Such models require data that is restructured in such a way that machines can understand e.g. by word embeddings or deep learning. Other methods of language modeling such as semantic networks (SN) can also structure the data without the need for training data. Apart from language modeling, KenSwQuAD can also benefit the language communities who can access this corpus that has texts of varying themes and sizes for their enjoyment and learning. The associated QA set can be used to test understanding of the language for reading enthusiasts who are learning the Swahili language.

This KenSwQuAD dataset of 1,445 Swahili text stories and its related collection of 7,526 QA pairs is the first of its kind for the low resource language of Swahili that employed our methodology. We annotated data from a corpus of narratives collected from Kenyan institutions of learning, publishers, and local information sources such as current affairs news stories. Our total number of questions is however lower than that of other high resource languages such as SQuAD [4], but this can be expected, based on our limited corpus that targeted only the texts collected from the Kencorpus project. We believe that annotating 56% of the whole of Kencorpus Swahili text corpus (87% of texts availed to annotators) was a good starting point for this gold standard set that is a first of its kind.

Challenges noted during annotation included the raw text being illegible. This was because the annotation was being done on the original primary data sources as collected from the field by the Kencorpus project [40]. The field data comprised of scanned documents such as PDF or image documents taken by phones as JPG or PNG. Some of these texts were unclear, either due to low resolution processing or inadequacy of the equipment that captured the images. Some handwritten texts were difficult to read, even when scanned properly. We advised the annotators to skip out any raw text that were unclear or illegible.

We missed out on several texts through skipping. There were about five skipped texts per annotator over the annotation project, though we had also done a purposive selection of texts to only get texts that were as clear as possible.

The initial intention of the research was to only do QA annotation on texts that had already been retyped to computer format, but these retyped texts were not yet available by the time of annotation, yet we were working within a tight timeframe. Preprocessing the images to text (TXT) format first then doing annotation thereafter would have been the preferred approach if this was possible. This would have dealt with issues such as improving the quality of the raw data, but probably not the issues of grammar or illegible handwriting by the original authors. Nonetheless, the corpus was a true reflection of the story texts from the field with the author's grammatical constructs being maintained. This could still be a good thing for research on natural language construct by authors and probably even the influence of language construct on machine learning. Of course, editing the original stories to provide better grammar is possible, though this would not be true to the original. This is also a possible research direction in creating a revised corpus which can then be used as a comparator with the original copy for linguistic and machine learning tasks. Typing out texts would also likely lead to the introduction of errors during the retyping, though quality control checks can be introduced to address this. Such issues also present a research opportunity for future work. The Kencorpus project has already presented retyped texts on the project website [40].

Other challenges were noted when dealing with the annotation process and tools. Some annotators indicated text data on numerical fields e.g. where we asked for paragraph numbers, the annotators would write the full text such as 'paragraph 3' instead of just '3'. This was easy to resolve at data cleaning stage. We also set the annotation data collection form to have all data fields as compulsory, ensuring that we were able to get data on each of the required fields. The issue of repeated story_IDs featured in 177 annotation cases. We however reconciled this set and come up with 75 unique texts from this set. Some texts were repeated but different QA pairs had been annotated regardless. This meant that we increased the number of QA pairs collected in some cases to be more than the standard 5 QA pairs per story text.

These repetitions likely occurred due to the methods of keeping track of work progress that the annotators employed. Some did not have an effective system to remind them of what they had already annotated, since all texts were posted in one location on each annotator's data repository. This issue mainly occurred during the initial weeks of the project. The errors became fewer with time as we continued to sensitize annotators on better methods of tracking work done, including moving such finalized works to different folders within their collections. However, we also lost 22 annotated texts (1.3% of annotations) whose story_IDs were captured incorrectly, and it was not possible to determine the correct story_IDs even after review. It would have taken lots of time and effort to determine the relationship between the QAs in this set and the correct story_ID from the collection of over 1,600 uniquely numbered texts. This was a small loss of potential QA pairs that could have otherwise gone into our dataset.

Our annotation was also based on raw data as collected from the field. We used texts collected from both lower and higher-level school students. Some of these texts had grammatical or logical errors. These rendered some texts difficult to follow, hence difficult to annotate with QA pairs that could make sense of the story text. We advised the annotators to skip any of such texts if they felt that setting 5 QA pairs was not possible. We had already resolved not to correct any errors on the raw text that was collected by the Kencorpus project. We were to remain true to source due to copyright and consent restrictions that the project had adopted.

Budget and time constraint meant that we could not undertake full annotation of all the texts on the Swahili dataset from Kencorpus. It would have been desirable to annotate as much of the available texts as possible, including the shorter or even longer texts, to benefit the researchers who are keen at testing machine learning models that need such data extremes. KenSwQuAD is however a good start, and more work can still be done on the unannotated part of the Kencorpus Swahili data. It is also possible to use our methodology to annotate any of the other two low-resource language texts (Dholuo, Luhya) in the Kencorpus collection. All the Kencorpus texts and datasets are available for free access and use by researchers on the project website [40].

# 7 CONCLUSION

This research developed the Kencorpus Swahili Question Answer Dataset, KenSwQuAD. This is a Swahili language question answer (QA) dataset of 7,526 QA pairs from a corpus of 1,445 story texts. Each text is annotated with at least 5 QA pairs. The dataset is the first of a kind employing our methodology that is created specifically for the low resource language of Swahili. Swahili, also known as Kiswahili, is a major language of communication in Eastern Africa and is also spoken in many other countries such as USA, Australia, and Europe [14]. This dataset is meant to spur interest in low resource languages for enthusiasts who may want to learn and test their understanding, as well as the members of the machine learning community, who may need a gold standard dataset to test their models in machine reading comprehension (MRC). The results of such modeling are practical end user systems such as internet search, dialog systems and chatbots, which can be done directly in the low resource languages such as Swahili, instead of relying on systems that are tailored for high resource languages such as English. Users prefer to use their usual languages of communication in natural language tasks and are usually just forced by lack of processing tools to switch to other languages [46].

The KenSwQuAD dataset was developed by human annotators who got their primary data from the larger Kencorpus project [40]. The Kencorpus Kenyan languages corpus collected data in three low resource languages of Africa, namely Swahili, Dholuo and Luhya. The collected Swahili data was both speech and text, with KenSwQuAD being created from 56% of the Swahili text stories in that project. This means that KenSwQuAD annotated more than half of the collected Swahili texts in the corpus. The annotators used a guide that spelt out all aspects that guided their work. A quality check mechanism was employed to confirm that the annotated questions were matching with the provided answers. Additionally, we built a proof-of-concept machine learning system based on semantic networks and a BERT-based model. These models confirmed that the QA set is capable of usage in a typical machine learning system as a gold standard set.

This research has provided insights on how to develop any typical QA dataset for a low resource language and should therefore be useful for research interest in QA annotation and dataset creation. Challenges in developing the QA dataset for the low resource language included the nature, format, and processing of the original raw texts themselves. In some cases, the authors of the original stories had not used language that was grammatically correct. The stories themselves did not make sense in some cases. However, such issues were resolved by exclusion criteria, though this was done at the expense of reducing the total number of texts available for annotation. The purposive sampling employed assisted in addressing such issues. The sampling was restrictive to particular types of texts i.e. short prose texts, hence longer texts that may probably have been useful in challenging machine learning systems were not considered. These are opportunities for further research.

This QA dataset of Swahili, KenSwQuAD, is now available as both a collection of 1,445 story texts and a separate comma separated values (CSV) file with the 7,526 QA pairs with at least 5 QA pairs per text. It is released as an open-source dataset to the research community for further review, research, and exploration. Future research is possible in areas such as expanding this QA dataset by inclusion of more story texts that were not annotated due to sampling, time, and budget constraint. Our research methodology can also be used to do QA annotation of texts of other low resource languages such as Dholuo and Luhya, which is already collected as part of the Kencorpus project, or any other low resource language that has a text data repository. Availing such Swahili text corpus and annotated QA datasets to the public domain shall continue to be of real value to researchers. Such datasets as KenSwQuAD also provide researchers with many opportunities to test their machine learning models on a gold standard QA dataset. KenSwQuAD is publicly available on the project website at either of the links below:

**kencorpus.maseno.ac.ke/kenswquad**
**https://doi.org/10.7910/DVN/OTL0LM**

# ACKNOWLEDGMENTS
This research was made possible by funding from Meridian Institute through Lacuna Fund under grant no. 0393-S-001
We acknowledge the inputs of the following research assistants who did the annotations: Alice Muchemi, Eric Magutu, Henry Masinde, Naomi Muthoni, Patrick Ndung'u and Rose Nyaboke

# REFERENCES

[1] J. Pennington, R. Socher, and C. D. Manning, "GloVe: Global Vectors for Word Representation," in EMNLP 2014 - 2014 Conference on Empirical Methods in Natural Language Processing, Proceedings of the Conference, pp. 1532–1543, 2014.

[2] J. Devlin, M.-W. Chang, K. Lee, and K. Toutanova, "Bert: Pre-Training of Deep Bidirectional Transformers for Language Understanding," arXiv preprint arXiv:1810.04805, 2018.

[3] J. Libovický, R. Rosa, and A. Fraser, "How Language-Neutral is Multilingual BERT?," arXiv preprint arXiv:1911.03310, 2019.

[4] P. Rajpurkar, J. Zhang, K. Lopyrev, and P. Liang, "Squad: 100,000+ Questions for Machine Comprehension of Text," arXiv preprint arXiv:1606.05250, 2016.

[5] M. Richardson, C. J. C. Burges, and E. Renshaw, "MCTest: A Challenge Dataset for The Open-Domain Machine Comprehension of Text," EMNLP 2013 - 2013 Conference on Empirical Methods in Natural Language Processing, Proceedings of the Conference, pp. 193–203, 2013.

[6] Y. Yang, W. T. Yih, and C. Meek, "WikiQA: A challenge dataset for open-domain question answering," in Proceedings of the 2015 Conference on Empirical Methods in Natural Language Processing, Sep. 2015, pp. 2013–2018.

[7] E. M. Voorhees and D. M. Tice, "Implementing a question answering evaluation," in Proceedings of LREC'2000 Workshop on Using Evaluation within HLT Programs: Results and Trends, May 2000.

[8] J. H. Clark et al., "TyDi QA: A Benchmark for Information-Seeking Question Answering in Typologically Diverse Languages," arXiv preprint arXiv:2003.05002, 2020.

[9] S. M. Yimam and M. Libsie, "TETEYEQ: Amharic Question Answering for Factoid Questions," IE-IR-LRL, vol. 3, no. 4, p. 17, 2009.

[10] L. Marais, "Approximating a Zulu GF Concrete Syntax with a Neural Network for Natural Language Understanding," 2021.

[11] J. E. Daniel, W. Brink, R. Eloff, and C. Copley, "Towards Automating Healthcare Question Answering in a Noisy Multilingual Low-Resource Setting," in Proceedings of the 57th Annual Meeting of the Association for Computational Linguistics, pp. 948–953, 2019.

[12] A. B. E. Mabrouk, M. B. H. Hmida, C. Fourati, H. Haddad, and A. Messaoudi, "A Multilingual African Embedding for FAQ Chatbots," arXiv preprint arXiv:2103.09185, 2021.

[13] W. S. Ismail and M. N. Homsi, "Dawqas: A Dataset for Arabic Why Question Answering System," Procedia Comput Sci, vol. 142, pp. 123–131, 2018.

[14] D. M. Eberhard, G. F. Simons, and C. D. Fennig, "Ethnologue: Languages of the World." SIL International, Dallas, TX, USA, 2021.

[15] wikipedia, "Swahili Language - Wikipedia," 2022. https://en.wikipedia.org/wiki/Swahili_language (accessed Jan. 20, 2022).

[16] omniglot, "Swahili Alphabet, Pronunciation and Language," 2021. https://omniglot.com/writing/swahili.htm (accessed Jan. 26, 2022).

[17] wikipedia, "Wikipedia." https://www.wikipedia.org/ (accessed Nov. 27, 2020).

[18] V. Berment, "Méthodes pour informatiser les langues et les groupes de langues peu dotées," Université Joseph-Fourier-Grenoble I, 2004.

[19] L. Besacier, E. Barnard, A. Karpov, and T. Schultz, "Automatic Speech Recognition for Under-Resourced Languages: A Survey," Elsevier B.V., 2014. doi: 10.1016/j.specom.2013.07.008.

[20] B. Wanjawa and L. Muchemi, "Model for Semantic Network Generation from Low Resource Languages as Applied to Question Answering–Case of Swahili," in 2021 IST-Africa Conference (IST-Africa), pp. 1–8, 2021.

[21] J. Hirschberg and C. D. Manning, "Advances in natural language processing," Science (1979), vol. 349, no. 6245, pp. 261–266, 2015.

[22] A. Hurskainen, "Helsinki corpus of Swahili," Compilers: Institute for Asian and African Studies (University of Helsinki) and CSC, 2004.

[23] aflat, "Kiswahili Part-of-Speech Tagger - Demo AfLaT.org," Sep. 2020. https://www.aflat.org/swatag

[24] K. Chege et al., Developing an Open source spell checker for Gĩkũyũ, Aflat, pp. 31, 2010.

[25] D. I. Adelani et al., "MasakhaNER: Named Entity Recognition for African Languages," arXiv preprint arXiv:2103.11811, 2021, Accessed: Jan. 29, 2022. [Online]. Available: https://arxiv.org/pdf/2103.11811

[26] S. M. Yimam and M. Libsie, "TETEYEQ: Amharic Question Answering for Factoid Questions," IE-IR-LRL, vol. 3, no. 4, p. 17, 2009.

[27] S. Muhie and M. Libsie, "Amharic Question Answering (AQA)," in 10th Dutch-Belgian Information Retrieval Workshop, 2010.

[28] T. A. Taffa and M. Libsie, "Amharic Question Answering for Biography, Definition, and Description Questions," in Proceedings of the 2019 Workshop on Widening NLP, Aug. 2019, pp. 110–113. [Online]. Available: https://aclanthology.org/W19-3635

[29] K. H. Amare, "Tigrigna Question Answering System for Factoid Questions," MSc. Thesis, Addis Ababa University, 2016.

[30] F. Faisal, S. Keshava, M. M. ibn Alam, and A. Anastasopoulos, "SD-QA: Spoken Dialectal Question Answering for the Real World." arXiv, 2021. doi: 10.48550/ARXIV.2109.12072.

[31] D. W. Oard and F. C. Gey, "The TREC 2002 Arabic/English CLIR Track.," in TREC, 2002.

[32] A. Aouichat and A. Guessoum, "Building TALAA-AFAQ, a Corpus of Arabic Factoid Question-Answers for a Question Answering System," in International Conference on Applications of Natural Language to Information Systems, pp. 380–386, 2017.

[33] A. Peñas et al., "Overview of QA4MRE at CLEF 2011: Question Answering for Machine Reading Evaluation," in CLEF (notebook papers/labs/workshop), pp. 1–20, 2011.

[34] V. Novák, S. Hartrumpf, and K. Hall, "Large-scale Semantic Networks: Annotation and Evaluation," in Proceedings of the Workshop on Semantic Evaluations: Recent Achievements and Future Directions (DEW '09), pp. 37–45, 2009.

[35] B. Wanjawa and L. Muchemi, "Using Semantic Networks for Question Answering-Case of Low-Resource Languages Such as Swahili," in International Conference on Applied Human Factors and Ergonomics, pp. 278–285, 2020.

[36] A. Singhal, "Introducing the Knowledge Graph: Things, Not Strings - Inside Search," vol. 2013, no. 7/22/2013. 2012. Accessed: Nov. 05, 2017. [Online]. Available: http://insidesearch.blogspot.com/2012/05/introducing-knowledge-graph-things-not.html

[37] R. Wang, C. Conrad, and S. Shah, "Using Set Cover to Optimize a Large-Scale Low Latency Distributed Graph," Proceedings of the 5th USENIX Workshop on Hot Topics in Cloud Computing, Jun., 2013.

[38] S. Sankar, S. Lassen, and M. Curtiss, "Under the Hood: Building out the infrastructure for Graph Search," 2013. http://www.facebook.com/notes/facebook-engineering/under-the-hood-building-out-the-infrastructure-for-graph-search/10151347573598920/ (accessed Nov. 06, 2017).

[39] A. Selamat and N. Akosu, "Word-length Algorithm for Language Identification of Under-Resourced Languages," Journal of King Saud University - Computer and Information Sciences, vol. 28, no. 4, pp. 457–469, 2016, doi: 10.1016/j.jksuci.2014.12.004.

[40] B. Wanjawa, L. Wanzare, F. Indede, O. McOnyango, L. Muchemi, and E. Ombui, "Kencorpus - Kenyan Languages Corpus," 2022. https://kencorpus.co.ke/ (accessed May 05, 2022).

[41] A. Conneau and G. Lample, "Cross-lingual Language Model Pretraining," in Advances in Neural Information Processing Systems, pp. 7059–7069, 2019.

[42] G. Martin, M. E. Mswahili, Y.-S. Jeong, and J. Young-Seob, "SwahBERT: Language Model of Swahili," in Proceedings of the 2022 Conference of the North American Chapter of the Association for Computational Linguistics: Human Language Technologies, pp. 303–313, 2022.

[43] "RDF Grapher." Nov. 2021. [Online]. Available: https://www.ldf.fi/service/rdf-grapher

[44] Pytext, "XLM-RoBERTa." https://pytext.readthedocs.io/en/master/xlm_r.html (accessed Nov. 15, 2022).

[45] Paperswithcode, "Question Answering on SQuAD2.0." https://paperswithcode.com/sota/question-answering-on-squad20 (accessed Nov. 14, 2022).

[46] I. Orife et al., "Masakhane–Machine Translation for Africa," arXiv preprint arXiv:2003.11529, 2020.